\documentclass[10pt,twocolumn,letterpaper]{article}

\usepackage[pagenumbers]{cvpr} 

\usepackage{graphicx}
\usepackage{amsmath}
\usepackage{amssymb}
\usepackage{booktabs}
\usepackage{algorithm}
\usepackage{algorithmic}

\usepackage{float}
\usepackage{color}
\usepackage{booktabs}
\usepackage{multirow}
\usepackage{mathrsfs}
\usepackage{graphics}
\usepackage{bm}
\usepackage{subfloat}

\usepackage[pagebackref,breaklinks,colorlinks]{hyperref}

\usepackage[capitalize]{cleveref}

\def\cvprPaperID{8733} 
\def\confName{CVPR}
\def\confYear{2022}

\begin{document}
\title{Cluster-based Contrastive Disentangling for Generalized Zero-Shot Learning}

\author{Yi Gao\\
Sichuan University\\
Chengdu 610065, P. R. China\\
{\tt\small gaoyi@stu.scu.edu.cn}
\and
Chenwei Tang\\
Sichuan University\\
Chengdu 610065, P. R. China\\
{\tt\small tangchenwei@scu.edu.cn}
\and
Jiancheng Lv\\
Sichuan University\\
Chengdu 610065, P. R. China\\
{\tt\small lvjiancheng@scu.edu.cn}
}
\maketitle

\begin{abstract}
    Generalized Zero-Shot Learning (GZSL) aims to recognize both seen and unseen classes by training only the seen classes, in which the instances of unseen classes tend to be biased towards the seen class. In this paper, we propose a Cluster-based Contrastive Disentangling (CCD) method to improve GZSL by alleviating the semantic gap and domain shift problems. Specifically, we first cluster the batch data to form several sets containing similar classes. Then, we disentangle the visual features into semantic-unspecific and semantic-matched variables, and further disentangle the semantic-matched variables into class-shared and class-unique variables according to the clustering results. The disentangled learning module with random swapping and semantic-visual alignment bridges the semantic gap. Moreover, we introduce contrastive learning on semantic-matched and class-unique variables to learn high intra-set and intra-class similarity, as well as inter-set and inter-class discriminability. Then, the generated visual features conform to the underlying characteristics of general images and have strong discriminative information, which alleviates the domain shift problem well. We evaluate our proposed method on four datasets and achieve state-of-the-art results in both conventional and generalized settings.
\end{abstract}

\section{Introduction}
\label{sec:intro}
Zero-Shot Learning (ZSL) with the ability to recognize instances of unseen classes has attracted a lot of attention since it was proposed in 2009 \cite{ali2009Describing}. Most existing ZSL methods are based on generative models, e.g., Generative Adversarial Nets (GAN) \cite{gan} and Variational Auto-Encoder (VAE) \cite{vae}. The generative models construct visual-semantic embedding space and transfer the knowledge from seen classes to unseen classes \cite{AWA2-gbu, bucher2017generating}. Then, the ZSL problem can be reformulated as fully-supervised classification tasks by generating visual features of unseen classes conditioned on the attribute descriptions \cite{xian2018feature,f-vaegan-d2}. However, the semantic gap between visual space and semantic space leads to a strong bias towards the trained seen classes, i.e., domain shift problem \cite{tang2021zero}. Especially in the Generalized ZSL (GZSL) setting, where both seen and unseen classes are used for the test, the domain shift problem is more serious \cite{free, SDG}. 

\begin{figure}[t]
  \centering
   \includegraphics[width=0.99\linewidth]{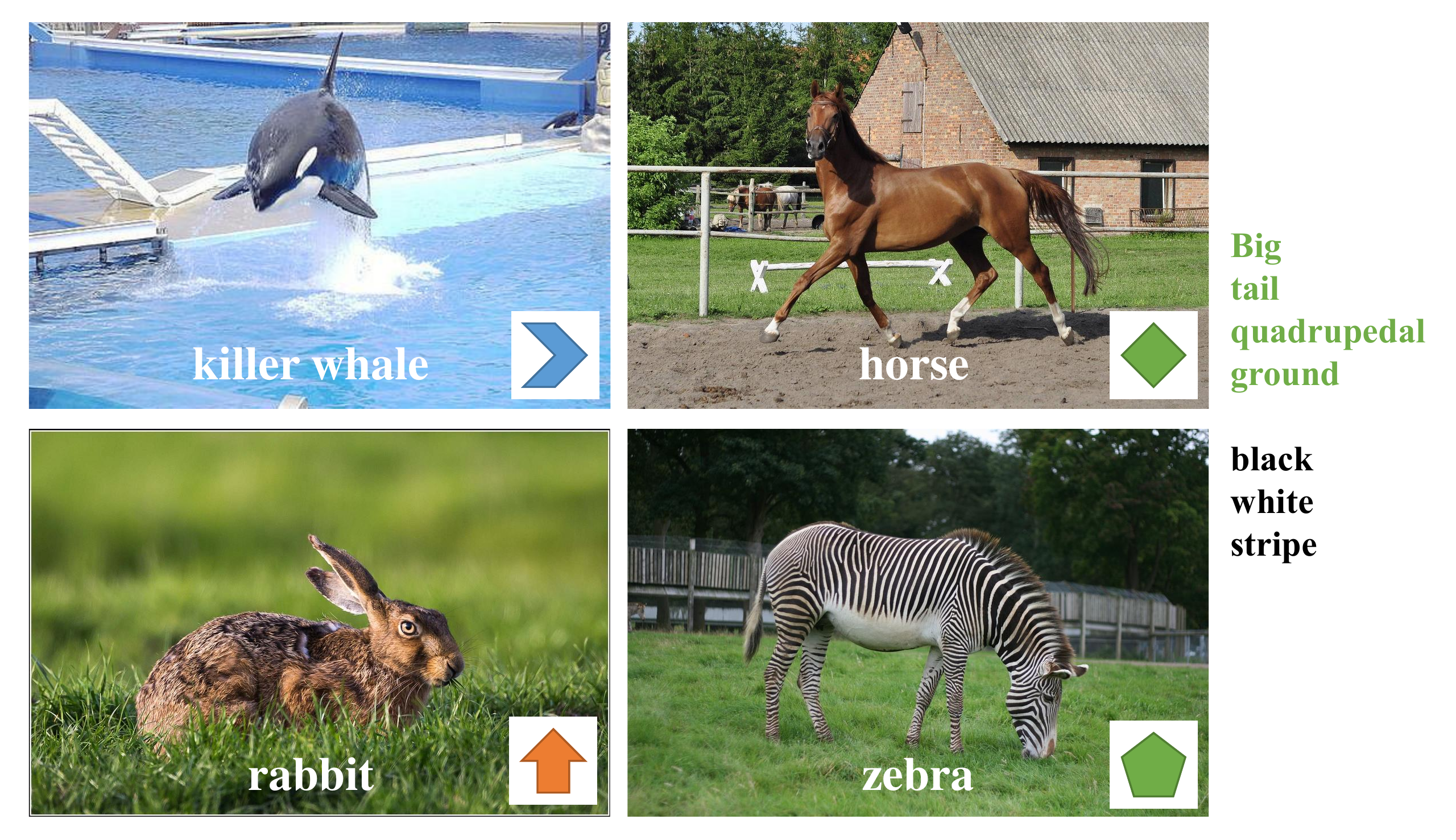}
   \caption{An example of humans imagining the instance of an unseen class. Suppose that the killer whale, horse, and rabbit are seen classes, and the zebra is the unseen class. The attributes of green font represent the class-shared attribute between zebra and horse, and the rests represent the class-unique attributes of zebra.}
   \label{fig:example}
\end{figure}

Actually, human beings do not directly imagine instances of novel categories only according to the semantic descriptions provided. As shown in \Cref{fig:example}, suppose that we have seen killer whales, horses, and rabbits. When we know the attribute descriptions (e.g., big, tail, quadrupedal, ground, black, white, and stripe) of the novel category of zebra, how do we imagine it? First, according to the attributes of big, tail, quadrupedal, ground, and etc., we can preliminarily judge that zebras and horses are similar in appearance. Then, we can roughly imagine zebras based on the \textit{class-shared attributes} between zebras and horses, referring to the same attributes between two classes, rather than the killer whales. After that, we can utilize the black and white stripes, and other \textit{class-unique attributes} to improve the appearance of zebras in our mind. In addition, we can also associate some features other than the given attribute descriptions, such as grass and short fur, so as to make our imagination more accurate.

Some recent ZSL works \cite{aaaigeneralized,SDG} introduce disentangled representation learning \cite{factorvae,dipvae,btcvae} to disentangle feature vectors into multiple independent variables. By the disentangled representations, e.g., environment variables and object variables, the latent vectors used to reconstruct images can be controlled as expectations \cite{aaaigeneralized}. Considering that all factors cannot be completely disentangled, the semantics disentangling method proposed in \cite{SDG} simply disentangles the visual feature vectors into two parts, i.e. semantic-consistent representation and semantic-unrelated representation. With the help of semantics disentangling representations, the domain shift problem is effectively alleviated. However, from the \Cref{fig:example}, we can observe that it is easier for humans to imagine novel instances based on similar categories (e.g. imagining zebras based on horses), but it is difficult to imagine instances based on far different categories (e.g. imagining zebras based on killer whales).  
 
Inspired by the process of human imagination, We propose to make full use of the latent representation of visual features, as well as the similarity and uniqueness between attribute descriptions. Then, we present a Cluster-based Contrastive Disentangling (CCD) method to improve GZSL by bridging the semantic gap and alleviating the domain shift problem. Specifically, we first design a clustering module to aggregate the batch samples containing $M$ classes into $N$ sets, where $N<<M$ and classes in a set are similar (e.g., horse and zebra belong to a set). Then, we introduce the disentangled representation learning module \cite{factorvae, aaaigeneralized,SDG} to disentangle the visual features into two parts. The first part is the semantic-matched features corresponding to the labeled attributes (e.g., big, ground, and stripe). Another is the semantic-unspecific features corresponding to the unspecified attributes (e.g., grass and short fur). Combined with the clustering results, we further disentangle the semantic-matched features into class-shared variables (e.g., visual features corresponding to big and ground attributes) and class-unique variables (e.g., visual features corresponding to black and white stripes attributes). Moreover, we introduce the contrastive learning \cite{bias1,bias2,bias3} to increase intra-set and intra-class similarity, as well as inter-set and inter-class discriminability \cite{tfvaegan,free,bias4,bias5}. 

Our contributes can be summarized as follows: (i) The disentangled learning module with random swapping and semantic-visual alignment module bridges the semantic gap by disentangling the visual latent vector into semantic-unspecific, class-shared, and class-unique representations. (ii) The contrastive learning module on cluster set and class-unique variables learns the semantic consistency, which alleviate the domain shift problem by increasing the intra-set and intra-class similarity, as well as inter-set and inter-class discriminability. (iii) Extensive experimental results on four benchmarks in both ZSL and GZSL show that our method can achieve superior or competitive performance compared with the state-of-the-art methods.

\section{Related Work}
\label{sec:related}
\subsection{Zero-Shot Learning}
Generative models, e.g., VAEs \cite{vae} and GANs \cite{gan}, shine in ZSL and GZSL. Most recent excellent ZSL methods use generative models \cite{vae1,CADA,epen,bucher2017generating,xian2018feature,f-vaegan-d2,tfvaegan,free,SDG,tang2021zero} to generate samples of unseen classes conditioned on semantic descriptions, and then transform ZSL into the fully supervised task. CADA-VAE \cite{CADA} learns the implicit space of images and semantics via aligned VAE. TF-VAEGAN \cite{tfvaegan} uses VAE and GAN to ensure semantic consistency for GZSL, which proposes a feedback module to constrain the expression of the generator. Compared with ZSL, the test set of GZSL contains both seen and unseen classes. An obvious problem is that the model will be biased towards the seen classes when inferring. FREE \cite{free} proposes a Feature Refinement (FR) module to refine the visual features of seen and unseen classes by integrating semantic → visual mappings into a unified generative model. In this paper, the proposed CCD method considers the semantic consistency of the generated samples, as well as emphasizes the reasonableness and discriminability of the generated samples.

\subsection{Disentangled Representation Learning}
Disentangled representation learning means decomposing the feature representation into multiple factors that are independent of each other \cite{bengio2013representation}, which is emulating the process of human cognition in the hope of learning some disentangled high-dimensional representations. There are many VAE-based disentangled representation learning, e.g., $\beta$-vae \cite{bvae} and factor-VAE \cite{factorvae}. ZSL needs to establish the mapping relationship between seen and unseen classes. Obviously, disentangling the latent vector is beneficial to an interpretable generation of unseen images. DRGZL \cite{aaaigeneralized} proposes dubbed disentangled-VAE, which decomposes visual and semantic features into category-distilling factors and category-dispersing factors. SDGZSL \cite{SDG} argues that the amount of information in images is greater than semantic description. In order to alleviate the semantic gap between the two modals, SDGZSL introduces a novel semantics disentangling framework, which factorizes the latent vectors obtained by encoding visual features into semantic-consistent and semantic-unrelated latent vectors. Notably, SDGZSL only uses semantic-consistent vectors when classifying. In this paper, we argue that the domain shift problem of generated images can be alleviated by sharing information between learning classes, including shared information between seen classes, and shared information between seen classes and unseen classes. Further, in order to improve the discriminability of features, it is also necessary to learn the discrimination information between classes.
\begin{figure*}[htbp]
  \centering
   \includegraphics[width=0.99\linewidth]{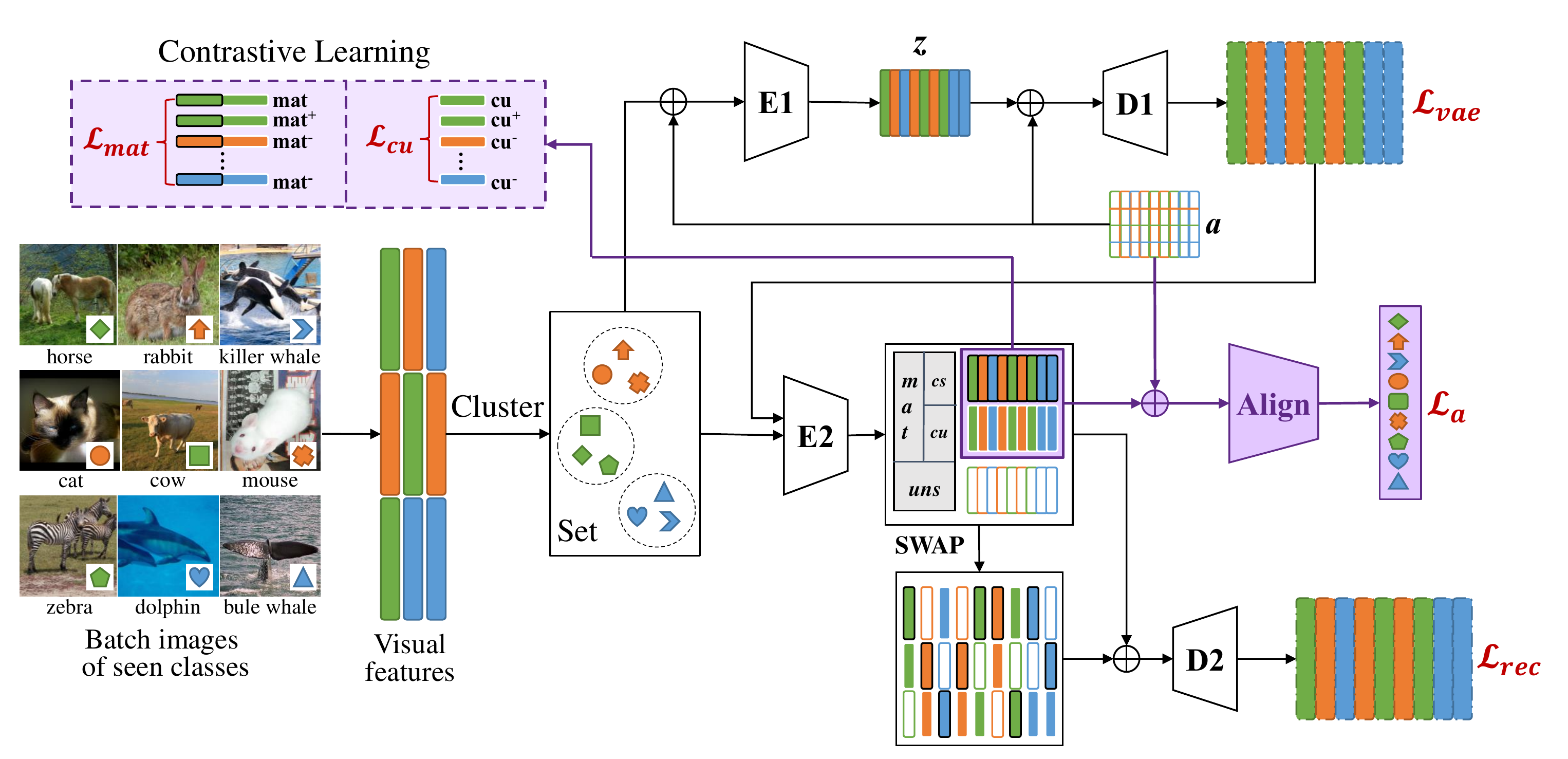}
   \caption{ An overall architecture of the Proposed Cluster-based Contrastive Disentangling (CCD) method. $mat$, $uns$, $cs$, and $cu$ denote the disentangled semantic-matched, semantic-unspecific, class-shared, and class-unique representations, respectively. $\oplus$ denotes vector concatenation.}
   \label{fig:model}
\end{figure*}

\subsection{Contrastive Learning}
Contrastive learning \cite{moco,simclr,mocov2} focuses on learning the common features between similar instances and distinguishing the differences between non-similar instances. Contrastive learning only needs to learn to distinguish the data in the feature space of the abstract semantic level, and its generalization ability is stronger. Recently, many contrastive learning methods \cite{simclrv2,mocov3} have been proposed and have refreshed the unsupervised classification accuracy of imagenet. Simclr \cite{simclr} and moco \cite{moco} augment an image as a positive sample, and other images as negative samples. The contrastive learning methods learn better representation by ensuring the distance between positive samples is much smaller than the distance between positive and negative samples. In this paper, we broaden the selection of positive samples, i.e., extend the augmentation of the same sample to similar samples (in the same cluster set).

\section{The Proposed CCD Method}
\label{sec:method}
In this section, we first introduce some notations and the definition of problem. We define $\mathcal{S}$ and $\mathcal{U}$ as seen classes with $N_s$ categories and unseen classes with $N_u$ categories, as well as $\mathcal{Y}_s$ and $\mathcal{Y}_u$ as their corresponding labels, respectively. Note that $\mathcal{Y}_s$ and $\mathcal{Y}_u$ have no intersection. We define $x_i$ as all visual features of class $i$, i.e., $\mathcal{S}=\{x_i^s, y_i^s\}_{i=1}^{N_{s}}$ and $\mathcal{U}=\{x_i^u, y_i^u\}_{i=1}^{N_{u}}$. In ZSL, we use $\mathcal{S}$ and semantic descriptions of seen classes $\mathcal{A}_s=\{a_i^s\}_{i=1}^{N_{s}}$ in training, and the test set only contains unseen classes $\mathcal{U}$. In GZSL, the test set contains both $\mathcal{S}$ and $\mathcal{U}$. The model learned in generative-based methods uses $a^{u}$ to synthesize $\hat{x}^{u}$, and then the samples of the test set and the synthesized samples are put into the classifier together.

\subsection{A Hybrid Twice-Generated GZSL Framework}
\label{twice-generated}
\Cref{fig:model} shows the overall architecture of our proposed CCD method for both ZSL and GZSL. We first sample a batch of instances of seen classes with $N$ categories, and cluster them into $M$ set. Then, we employ VAE and Auto-Encoder (AE) to generate the visual features and disentangled visual features conditioned on the attributes descriptions. 

For the first generative model based on VAE, we denote the distributions of encoder (the E1 in \Cref{fig:model}) and decoder (the D1 in \Cref{fig:model}) as $q_{\phi}(z|x,a)$ and $p_{\theta}(x|z,a)$, respectively. The $p_{\theta}(z|a)$ denotes the prior distribution, which is usually chosen as a Gaussian distribution, i.e., $p_{\theta}(z|a) \sim \mathcal{N}(0,1)$. The encoder (E1) distrubution $q_{\phi}(z|x,a)$ is used to approximate $p_{\theta}(z|a)$. As in the standard VAE, the loss $\mathcal{L}_{vae}$ is defined as follows:
\begin{align}
\begin{split}
    \mathcal{L}_{vae} & =\mathbb{E}_{q_{\phi}(z|x,a)}\left[\log p_{\theta}(x|z,a)\right] \\ 
    & -D_{K L}\left(q_{\phi}(z |x,a) \| p_{\theta}(z|a)\right),
\end{split}
\label{eq:vaeloss}
\end{align}
where the first item is the reconstruction loss, and the second item is the Kullback-Leibler divergence between $q_{\phi}(z|x,a)$ and $p_{\theta}(z|a)$.

For the second generative model based on AE, we use the encoder (the E2 in \Cref{fig:model}) to disentangle the latent visual features in clustering set into semantic-matched variables ($mat$) and semantic-unspecific variables ($uns$), i.e., $E2: x \in \mathbb{R}^{d} \rightarrow mat \in \mathbb{R}^{mat}, uns \in \mathbb{R}^{uns}$. Then, the semantic-matched variables ($mat$) are further disentangled into class-shared variable ($cs$) and class-unique variable ($cu$), i.e., $E2: mat \in \mathbb{R}^{mat} \rightarrow cs \in \mathbb{R}^{cs}, cu \in \mathbb{R}^{cu}$. It is worth mention that, the pseudo visual features generated by the VAE will be also input to the AE for disentangled learning. Then, we randomly swap the semantic-unspecific variables ($uns$), class-shared variable ($cs$) and class-unique variable ($cu$), denote as $\overline{uns, cs, cu}$, for better disentangling representations. The decoder (the D2 in \Cref{fig:model}) reconstructs the visual features conditioned on both original disentangling representations and swapped disentangling representations. The reconstruction loss $\mathcal{L}_{rec}$ can be given by
\begin{equation}
\mathcal{L}_{rec}=\sum_{{x} \in \mathcal{S}}\left\|{x}-D2((mat,uns), \overline{uns, cs, cu}) \right\|^{2},
\label{eq:rec}
\end{equation}
where $\mathcal{S}$ refers to all seen class samples in the training set. The $(mat,uns)$ and $\overline{uns, cs, cu}$ denotes the original disentangling representations and swapped disentangling representations, respectively. 

\subsection{Set-based \& Class-based Contrastive Learning}
\label{con-U}
In nature, most objects have some similarities, and we assume that there are also specialized nerves in the human brain that preserve this general underlying similarity information. Inspired by this, we argue that it is more appropriate to learn the underlying similarity information among similar objects. Compared to the class level, We propose set-based contrastive learning, which learns more general information and helps to improve the generalization of the model. Here we choose a clustering algorithm k-means\cite{kmeans} to facilitate the construction of similar samples in a batch. We aggregate a batch of samples into $M$ sets and then modify their labels to $0 - (M-1)$. The number of samples in each set is  $\left\{s_{0},s_{1},...s_{M-1}\right\}$. It is natural to think that the network can be guided to learn the underlying similarity information within set. We treat samples from the same set as positive samples and those from different sets as negative samples. Set-based contrast learning allows to increase the similarity of samples within the set and reduce the similarity between sets. The loss function of the positive and negative samples between the clustering sets is as follows:
\begin{equation}
\label{eq:mat}
\mathcal{L}_{mat}=-\log \frac{\exp \left(mat \cdot mat^{+} / \tau\right)}{\sum_{i=0}^{M-1} \sum_{j=0}^{s_i} \exp \left(mat \cdot mat_{ij} / \tau\right)},
\end{equation}
where $mat$, $mat^+$, and $mat_{ij}$ respectively define the semantic-matched vector, a positive sample semantic-matched vector similar to semantic-matched, and the j-th semantic-matched vector of the i-th set. $\tau$ is the temperature parameter. By pushing the distance between the sets farther, model can learn more general discrimination information than the category level. More importantly, by narrowing the distance within set, model can learn high-level abstract representations of similar features, which is conducive to strengthening the stability of the model generated samples and preventing domain shift problems.

Considering that one clustering set contains multiple categories, the discriminative information is still coarse-grained. To improve the discriminative information content of class-unique disentangled variable ($cu$), we feed each of the M sets into the trainer, and use class-based contrastive learning to learn discriminative information between similar classes. Suppose there are $K$ classes in a set. The number of samples in each class is $\left\{c_{0}, c_{1}, ..., c_{K}\right\}$. We construct positive and negative samples according to the following strategy. The samples of the same category in a set are mutually positive samples, and the samples of different categories are mutually negative samples. Similarly, The loss function in a set is as follows:
\begin{equation}
\label{eq:cu}
\mathcal{L}_{cu}=-\log \frac{\exp \left(cu \cdot cu^{+} / \tau\right)}{\sum_{i=0}^{K-1} \sum_{j=0}^{c_i} \exp \left(cu \cdot cu_{i,j} / \tau\right)},
\end{equation}
where $cu$, $cu^+$, and $cu_{i}$ respectively define the class-unique vector, a positive sample class-unique vector similar to $cu$, and the j-th class-unique vector of the i-th class. $\tau$ is the temperature parameter. Contrastive learning in the set can explicitly learn the differences between each class, making $cu$ more discriminative.

\subsection{Semantic Alignment Disentangling}
\label{align}
We use a random swap strategy to guarantee the disentangling of semantic-matched variables ($mat$) and semantic-unspecific variables ($uns$). In a batch of training, we denote the latent vector encoded by encoder E2 of AE as ${z}=\left[{uns}, {mat}\right]$. We have 
 \begin{equation}
\left\{{Z}_{i}\right\}_{i=1}^{B}=\left\{{uns}_{i},{mat}_{i}\right\}_{i=1}^{B},
\end{equation}
and
\begin{equation}
\left\{{mat}_{i}\right\}_{i=1}^{B}=\left\{{cs}_{i},{cu}_{i}\right\}_{i=1}^{B}.
\end{equation}
Then we generate a set of random indexes with the number of batch-size, and define it as index. We have
 \begin{equation}
\left\{{Z}_{i}^{\prime}\right\}_{i=1}^{B}=\left\{{uns}_{index[i]},{mat}_{i}\right\}_{i=1}^{B},
\end{equation}
and
\begin{equation}
\left\{{Z}_{i}^{\prime\prime}\right\}_{i=1}^{B}=\left\{{uns}_i, {cs}_{index[i]}, {cu}_i\right\}_{i=1}^{B}.
\end{equation}
Finally, we feed the randomly swapped ${Z}$, ${Z^{\prime}}$ and ${Z^{\prime\prime}}$ into AE's decoder D2 and use \cref{eq:rec} for training. 

To align semantic-matched variables ($mat$) with the semantic description, we designed a semantic alignment module that constrains that semantic-matched variables ($mat$) must be paired with the corresponding semantic description. The semantic alignment module learns the relationship between latent vectors and semantic descriptions, and can exclude the influence of non-semantics in subsequent classification. The loss function to optimize the alignment loss $\mathcal{L}_{a}$ is as follows:
\begin{equation}
\label{eq:a}
\mathcal{L}_{a}=-\frac{1}{BS} \sum_{i=0}^{BS-1} \sum_{c=0}^{BK-1} y_{i c} \log \left(p_{i c}\right),
\end{equation}
where $BS$ and $BK$ denote batch size and the number of categories in a batch, respectively.

Thus, the total loss of our method is formulated as:
\begin{equation}
\mathcal{L}_{total} = \mathcal{L}_{vae}+\mathcal{L}_{rec}+\alpha \mathcal{L}_{mat}+\beta \mathcal{L}_{cu}+\gamma \mathcal{L}_{a},
\end{equation}
where $\alpha$, $\beta$, and $\gamma$ are hyper-parameter. It is worth noting that the hyper-parameter $\alpha$, $\beta$, and $\gamma$ for different loss terms are
set via cross validation. Algorithm \ref{ouralgorithm} outlines the training procedures of the proposed CCD method.

\begin{algorithm}[ht]
  \caption{Training procedures of the proposed method.}
  \label{ouralgorithm}
  \begin{algorithmic}[1]
    \STATE {\bfseries Input:} \\
    \quad  $\mathcal{S}$: the training dataset of seen classes, \\
    \quad  $\mathcal{A}_s$: the attributes of seen classes,\\
    \quad  $N_{step}$, $B$: the training steps and batch size,\\
    \quad  $\alpha$, $\beta$, $\gamma$: the hyper-parameters.\\
    
    \STATE {\bfseries Initialize:} \\
    The parameters of E1, D1, E2, D2, and Align module are $\phi, \theta, \omega, \varphi, \psi$, respectively. 
    
    \FOR{$it=1,...,N_{step}$}
    \STATE Sample examples $\left\{{x}_{i}, {y}_{i}, {a}_{i}\right\}_{i=1}^{B}$ from $\mathcal{S}, \mathcal{A}_s$.
    \STATE Cluster into $M$ set, containing $M_0, M_1,...M_M$ classes, respectively.
     \STATE $z \gets E1(x,a)$, $\hat{x} \gets D1(z,a)$.
     \STATE Compute $\mathcal{L}_{vae}$ by \cref{eq:vaeloss}.
     \STATE Disentangle $x$ and $\hat{x}$ into $uns, mat(cs, cu)$ with E2.
    \FOR{$t$ steps}
     \STATE   Compute $\mathcal{L}_{a}$ and update $\psi$ by \cref{eq:a}.
    \ENDFOR
     \STATE $x^{\prime} \gets D2(uns, mat)$.
    \STATE Randomly swap $uns$, $cs$, and $cu$ for $\overline{uns, cs, cu}$.
    \STATE $x^{\prime\prime} \gets D2(\overline{uns, cs, cu})$, compute $L_{rec}$ by \cref{eq:rec}.
    \STATE Compute $\mathcal{L}_{mat}$ between sets by \cref{eq:mat}.
    \FOR{$M$ steps}
     \STATE   Compute $\mathcal{L}_{cu}$ within set by \cref{eq:cu}.
    \ENDFOR
    \STATE   $\mathcal{L}_{total}=\mathcal{L}_{vae}+\mathcal{L}_{rec}+\alpha \mathcal{L}_{mat}+\beta \mathcal{L}_{cu}+\gamma \mathcal{L}_{a}$.
    \STATE Update $\phi, \theta, \omega, \varphi$.
    \ENDFOR
  \end{algorithmic}
\end{algorithm} 

\begin{table*}[ht]
\centering
\caption{Comparison with state-of-the-arts on four datasets. The top two results of the U, S and H are highlighted in bold. GEN and NON-GEN refer to generative and non-generative methods, i.e., embedding-based methods. * means the fine-tuned backbone is used.}

\setlength{\tabcolsep}{1.6mm}{
\begin{tabular}{c|c|ccc|ccc|ccc|ccc} 
\toprule
\multicolumn{2}{c|}{\multirow{2}{*}{methods}} & \multicolumn{3}{c|}{CUB}                      & \multicolumn{3}{c|}{AWA2}                                                                         & \multicolumn{3}{c|}{SUN}                                                                          & \multicolumn{3}{c}{FLO}                                                                           \\ 
\cline{3-14}
\multicolumn{2}{c|}{}                         & U             & S             & H             & U                                 & S                        & H                                  & U                        & S                                 & H                                  & U                        & S                                 & H                                  \\ 
\midrule
\multirow{7}{*}{NON-GEN} & TCN(2019)\cite{tcn}          & 52.6          & 52.0          & 52.3          & 61.2                              & 65.8                     & 63.4                               & 31.2                     & 37.3                              & 34.0                               & -                        & -                                 & -                                  \\
                         & DVBE(2020)\cite{DVBE}         & 53.2          & 60.2          & 56.5          & 63.6                              & 70.8                     & 67                                 & 45.0                     & 37.2                              & 40.7                               & -                        & -                                 & -                                  \\
                         & DVBE*(2020)\cite{DVBE}         & 64.4          & 73.2          & 68.5          & 62.7                              & 77.5                     & 69.4                               & 44.1                     & 41.6                              & 42.8                               & -                        & -                                 & -                                  \\
                         & RGEN(2020)\cite{RGEN}          & 60.0          & 73.5          & 66.1          & 67.1                              & 76.5                     & 71.5                               & 44.0                     & 31.7                              & 36.8                               & -                        & -                                 & -                                  \\
                         & APN(2020)\cite{APN}           & 65.3          & 69.3          & 67.2          & 56.5                              & 78.0                     & 65.5                               & 41.9                     & 34.0                              & 37.6                               & -                        & -                                 & -                                  \\
                         & GEM-ZSL*(2021)\cite{goal}     & 64.8          & 77.1          & 70.4          & 64.8                              & 77.5                     & 70.6                               & 38.1                     & 35.7                              & 36.9                               & -                        & -                                 & -                                  \\
                         & DCEN*(2021)\cite{bias3}        & 63.8          & 78.4          & 70.4          & 62.4                              & 81.7                     & 70.8                               & 43.7                     & 39.8                              & 41.7                               & -                        & -                                 & -                                  \\ 
                        & DPPN*(2021)\cite{DPPN}        & 70.2          & 77.1          & 73.5          & 63.1                              & 86.8                     & 73.1                               & 47.9                     & 35.8                              & 41.0                               & -                        & -                                 & -                                  \\ 
\midrule
\multirow{14}{*}{GEN}    & f-VAEGAN-D2(2019)\cite{f-vaegan-d2}  & 48.4          & 60.1          & 53.6          & 57.6                              & 70.6                     & 63.5                               & 45.1                     & 38.0                              & 41.3                               & 56.8                     & 74.9                              & 64.6                               \\
                         & f-VAEGAN-D2*(2019)\cite{f-vaegan-d2} & 63.2          & 75.6          & 68.9          & 57.1                              & 76.1                     & 65.2                               & \textbf{50.1}            & 37.8                              & \textbf{43.1 }                     & 63.3                     & 92.4                              & 75.1                               \\
                         & TF-VAEGAN(2020)\cite{tfvaegan}    & 52.8          & 64.7          & 58.1          & 59.8                              & 75.1                     & 66.6                               & 45.6                     & 40.7                              & 43.0                               & 62.5                     & 84.1                              & 71.7                               \\
                         & TF-VAEGAN*(2020)\cite{tfvaegan}   & 63.8          & \textbf{79.3} & 70.7          & 55.5                              & \textbf{83.6}            & 66.7                               & -                        & -                                 & -                                  & 69.5                     & 92.5                              & 79.4                               \\
                         & E-PGN(2020)\cite{epen}        & 52.0          & 61.1          & 56.2          & 52.6                              & \textbf{83.5}            & 64.4                               & -                        & -                                 & -                                  & 71.5                     & 82.2                              & 76.5                               \\
                         & LsrGAN(2020)\cite{lsrgan}       & 48.1          & 59.1          & 53.0          & 54.6                              & 74.6                     & 63.0                               & 44.8                     & 37.7                              & 40.9                               & -                        & -                                 & -                                  \\
                         & DDV(2021)\cite{aaaigeneralized}          & 51.1          & 58.2          & 54.4          & 56.9                              & 80.2                     & 66.6                               & 36.6                     & \textbf{47.6}                     & 41.4                               & -                        & -                                 & -                                  \\
                         & GCM-CF(2021)\cite{gcm}       & 61.0          & 59.7          & 60.3          & 60.4                              & 75.1                     & 67.0                               & \textbf{47.9 }           & 37.8                              & 42.2                               & -                        & -                                 & -                                  \\
                         & SDGZSL(2021)\cite{SDG}       & 59.9          & 66.4          & 63.0          & 64.6                              & 73.6                     & 68.8                               & -                        & -                                 & -                                  & 83.3                     & 90.2                              & 86.6                               \\
                         & SDGZSL*(2021)\cite{SDG}      & \textbf{73.0} & 77.5          & \textbf{75.1} & \textbf{69.6}                     & 78.2                     & \textbf{73.7}                      & -                        & -                                 & -                                  & \textbf{86.1}            & 89.1                              & \textbf{87.8}                      \\
                         & FREE(2021)\cite{free}         & 55.7          & 59.9          & 57.7          & 60.4                              & 75.4                     & 67.1                               & 47.4                     & 37.2                              & 41.7                               & 67.4                     & 84.5                              & 75.0                               \\
                         & GSMFlow(2021)\cite{gen1}      & 61.4          & 67.4          & 64.3          & 64.5                              & 82.1                     & 72.3                               & -                        & -                                 & -                                  & \textbf{86.6}            & 87.8                              & 87.2                               \\
                         & ours(CCD)          & 62.3          & 66.8          & 64.5          & 65.1                              & 73.4                     & 69.0                               & \multicolumn{1}{l}{40.2} & \multicolumn{1}{l}{44.4}          & \multicolumn{1}{l|}{42.1}          & 83.6                     & \textbf{92.7}                     & 87.9                               \\
                         & ours(CCD)*         & \textbf{73.1} & \textbf{78.9} & \textbf{75.9} & \multicolumn{1}{l}{\textbf{68.0}} & \multicolumn{1}{l}{82.7} & \multicolumn{1}{l|}{\textbf{74.6}} & \multicolumn{1}{l}{43.9} & \multicolumn{1}{l}{\textbf{44.6}} & \multicolumn{1}{l|}{\textbf{44.2}} & \multicolumn{1}{l}{82.4} & \multicolumn{1}{l}{\textbf{94.5}} & \multicolumn{1}{l}{\textbf{88.0}}  \\
\bottomrule
\end{tabular}}
\label{tab:allresults}
\end{table*}
\section{Experiments}
\label{sec:experiments}
\noindent\textbf{Datasets.}
We conduct full experiments on four benchmark datasets widely used for GZSL, i.e., CUB (Caltech UCSD Birds 200) \cite{CUB}, AWA2 (Animals with Attributes 2) \cite{AWA2-gbu}, SUN (SUN Attribute) \cite{patterson2012sun}, and FLO (Oxford Flowers) \cite{FLO}. The CUB and FLO datasets consist of 200 and 102 different categories, respectively. Each image of CUB and FLO datasets has 10 sentences of semantic descriptions, and we extract the 1024-dimensional semantic features from these descriptions by character-based CNN-RNN \cite{reed2016learning}. The AWA2 and SUN datasets consist of 50 and 717 categories, respectively. Each category of AWA2 and SUN is labeled with 85 and 102 attributes, respectively. We adopt the seen/unseen split setting proposed in \cite{AWA2-gbu}. All visual features are 2048-dimensional vectors extracted by 101-layered ResNet \cite{He2016Deep}.

\noindent\textbf{Evaluation Metrics.}
In GZSL, we will evaluate the accuracy of the seen ($\mathcal{S}$) and unseen classes ($\mathcal{U}$) in the test set, and then use their harmonic mean ($\mathcal{H}$) to evaluate the performance of the method, which can be formulated as $H = 2 * \mathcal{S} * \mathcal{U}/(\mathcal{S} + \mathcal{U})$.

\noindent\textbf{Implementation Details.}
The encoder and decoder of both VAE and AE are implemented using Multi-Layer Perceptron (MLP). The latent vectors of AE are decomposed into $uns$, $cs$, $cu$, and their dimensions are the same. The alignment module contains two pairs of FC+ReLU layers, and the dimension of the hidden layer is set as 4096. Our method is implemented with PyTorch and optimized by ADAM optimizer\cite{adam}. Following the general settings, the learning rate and temperature parameter is set to 3e-4 and 0.1, respectively. For the hyper-parameter $\alpha$, $\beta$, and $\gamma$, we set them as 0.1, 0.2, 2, respectively. \footnote{The source code will be released upon acceptance.}

\subsection{Comparison with State-of-the-Arts}
We compare our results with other state-of-the-art GZSL methods in \Cref{tab:allresults}, and our proposed method outperforms all of them. Some previous works also reported results of fine-tuning the backbone network on seen classes without violating ZSL regulations. Thus, we report the results of both fine-tuning and results without fine-tuning. We achieve state-of-the-art results on the CUB, AWA, and FLO datasets without fine-tuning, and achieve very competitive results on the SUN dataset. For the backbone network after fine-tuning, we surpass all comparison methods. Our proposed CCD method is inspired by the process of human imagination, i.e., making full use of the latent representation of visual features, as well as the similarity and uniqueness between attribute descriptions. The core idea corresponds to the cluster-based disentangling learning and contrastive learning between sets and classes. The experimental results prove that our idea is credible, and the proposed method is able to improve GZSL by alleviating the semantic gap and domain shift problems. Moreover, the proposed CCD method is significantly better than SDGZSL\cite{SDG}, DDV\cite{aaaigeneralized} and other GZSL methods based on disentangling learning.

\subsection{Conventional Zero-shot Learning Results}
Results in \Cref{tab:allresults} prove the effectiveness of our proposed method by achieving state-of-the-art results on most datasets in GZSL. Theoretically speaking, the proposed method improves the GZSL by alleviating the semantic gap and domain shift problem, which can also improve the performance of conventional ZSL to a certain extent. As \Cref{tab:zsl} shown, our proposed CCD method performs best performances on all four datasets in conventional ZSL setting. The comparison results further indicate that our proposed method can effectively alleviate the semantic gap and domain shift problem.

\begin{table}[ht]
\caption{Comparison with both conventional and generalized zero-shot learning methods, * means a fine-tuned backbone is used.}.
\begin{tabular}{@{}ccccc@{}}
\toprule
Method & CUB & AWA2 & SUN & FLO \\ \midrule
DCN(2018)\cite{DCN}       &  56.2  &  65.2     &  61.8   &  -   \\
TCN(2019)\cite{tcn}       &  59.5  &  71.2     &  61.5   &  -   \\
f-VAEGAN-D2(2019)\cite{f-vaegan-d2}       & 61.0    &   71.1   &  -   &    67.7 \\
   LsrGAN(2020)\cite{lsrgan}    &  60.3   &   66.4   &  62.5   &  -   \\
TF-VAEGAN(2020)\cite{tfvaegan}       &   64.9  &  72.2    &  -   &   70.8  \\
 TF-VAEGAN*(2020)\cite{tfvaegan}      &   74.3  &   73.4   &  -   &  74.7   \\
 APN(2020)\cite{APN}       &  72.0   &   68.4   &  61.6   &   -  \\
E-PGN(2020)\cite{epen}  &72.4 & 73.4     &  -   &   85.7  \\
 OCD-CAVE(2020)\cite{OCD-CVAE}    &  60.3   &   71.3   &   63.5  &  -   \\
   GEM-ZSL(2021)\cite{goal}     &  77.8   &   67.3   &  62.8   &   -  \\
SDGZSL(2021)\cite{SDG}       &  75.5   &  72.1    &  -   &   85.4  \\
SDGZSL*(2021)\cite{SDG}       &  78.5   &  74.3    &  -   &   86.9  \\
GSMFlow(2021)\cite{gen1}       &  76.4   &   72.7   &   -  &  86.9   \\
ours(CCD)    &  77.1   &  72.1    &   63.9  &  88.2    \\ 
ours(CCD)*   &  80.3   &  76.7    &   64.5  &  90.1  \\\bottomrule
\end{tabular}
\label{tab:zsl}
\end{table}

\subsection{Ablation Study}
\subsubsection{Component Analysis}
To verify the effectiveness of our proposed method, we performed a performance analysis on each component. \Cref{tab:component} shows the results of our various components, where $SWAP$ introduced in \cref{align} represents the random swapping strategy, and its role is to decouple $uns$ and $mat$. $ALIGN$ introduced in \cref{align} stands for semantic-visual alignment module, and it restricts mat to be consistent with the semantic description. $Con$-$U$ represents the method described in \cref{con-U}, which is the contrastive learning between sets, and $Con$-$D$ represents the method described in \cref{con-U}, which is the contrastive learning within the set. Their role is to disentangle $cs$ and $cu$ respectively. The performance of the model has been greatly improved after the inter-set and inter-class contrastive learning, which proves that the model has learned strong discriminative information. In addition, the ALIGN module guarantees the consistency of semantics, and the SWAP strengthens this point, which fully decouples $uns$ and $mat$.

\begin{table}[t]
\caption{Component effectiveness analysis on CUB and AWA2 datasets. Baseline is the hybrid twice-generated GZSL framework.}
\begin{tabular}{@{}l|lll|lll@{}}
\toprule
\multicolumn{1}{c|}{\multirow{2}{*}{methods}}                                          & \multicolumn{3}{c|}{CUB}                             & \multicolumn{3}{c}{AWA}             \\ \cmidrule(l){2-7} 
\multicolumn{1}{c|}{}                                                                  & \multicolumn{1}{c}{U} & \multicolumn{1}{c}{S} & H    & \multicolumn{1}{c}{U} & S    & H    \\ \midrule
baseline                                                                               & 67.7                  & 70.2                  & 68.9 & 60.4                  & 71.8 & 65.6 \\
\ \ $+ALIGN $                                                            & 69.2                  & 71.8                  & 70.5 & 62.3                  & 76.2 & 68.6 \\
\ \ \ \ $+SWAP$                                         & 69.7                  & 73.2                  & 71.4 & 64.1                  & 78.5 & 70.6 \\
\ \ \ \ \ \ $+Con$-$U$                   & 70.4                  & 75.3                  & 72.8 & 64.6                  & 79.4 & 71.2 \\
\ \ \ \ \ \ \ \ $+Con$-$D$ & 73.1                  & 78.9                  & 75.9 & 68.0                  & 82.7 & 74.6 \\ \bottomrule
\end{tabular}
\label{tab:component}
\end{table}

\begin{figure}[t]
  \centering
   \includegraphics[width=0.99\linewidth]{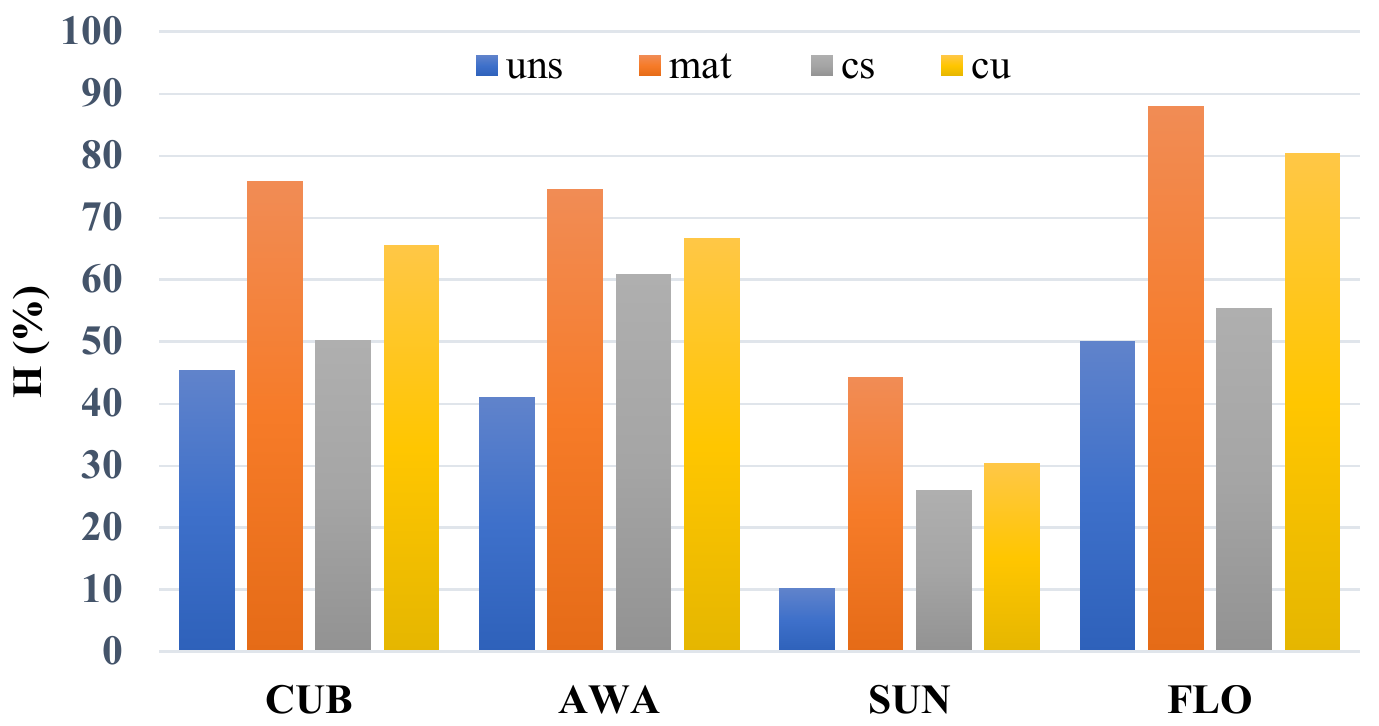}
   \caption{The effect of disentangling learning. We use $uns$, $mat$, $cs$, and $cu$ to train the classifiers, respectively.}
   \label{disentangling}
\end{figure}

\subsubsection{Effectiveness Analysis of Disentangling}
To prove that our disentangling method works, we need to verify the effectiveness of $uns$, $mat(cs+cu)$, $cs$, and $cu$ respectively. The final accuracy of the paper is classified using mat. Then we use the other three disentangled vectors to train the classifier. Intuitively, uns's results will be the lowest, because it encodes parts of the semantic description that cannot be aligned with visual features. mat will be the highest because it contains both basic information and discriminative information. Although $cu$ contains strong discriminant information, the lack of basic information can lead to misjudgment, so he is the second place. Since the remaining $cs$ contains basic information shared by the class, note that this information is relative to the set, and its accuracy will be lower than that of $cs$, it is not easy for the classifier to determine which class in the set test sample belongs to. \Cref{disentangling} corroborates our intuition.

\begin{figure}[t]
  \centering
  \begin{subfigure}{0.49\linewidth}
  \centering
    \includegraphics[width=4cm]{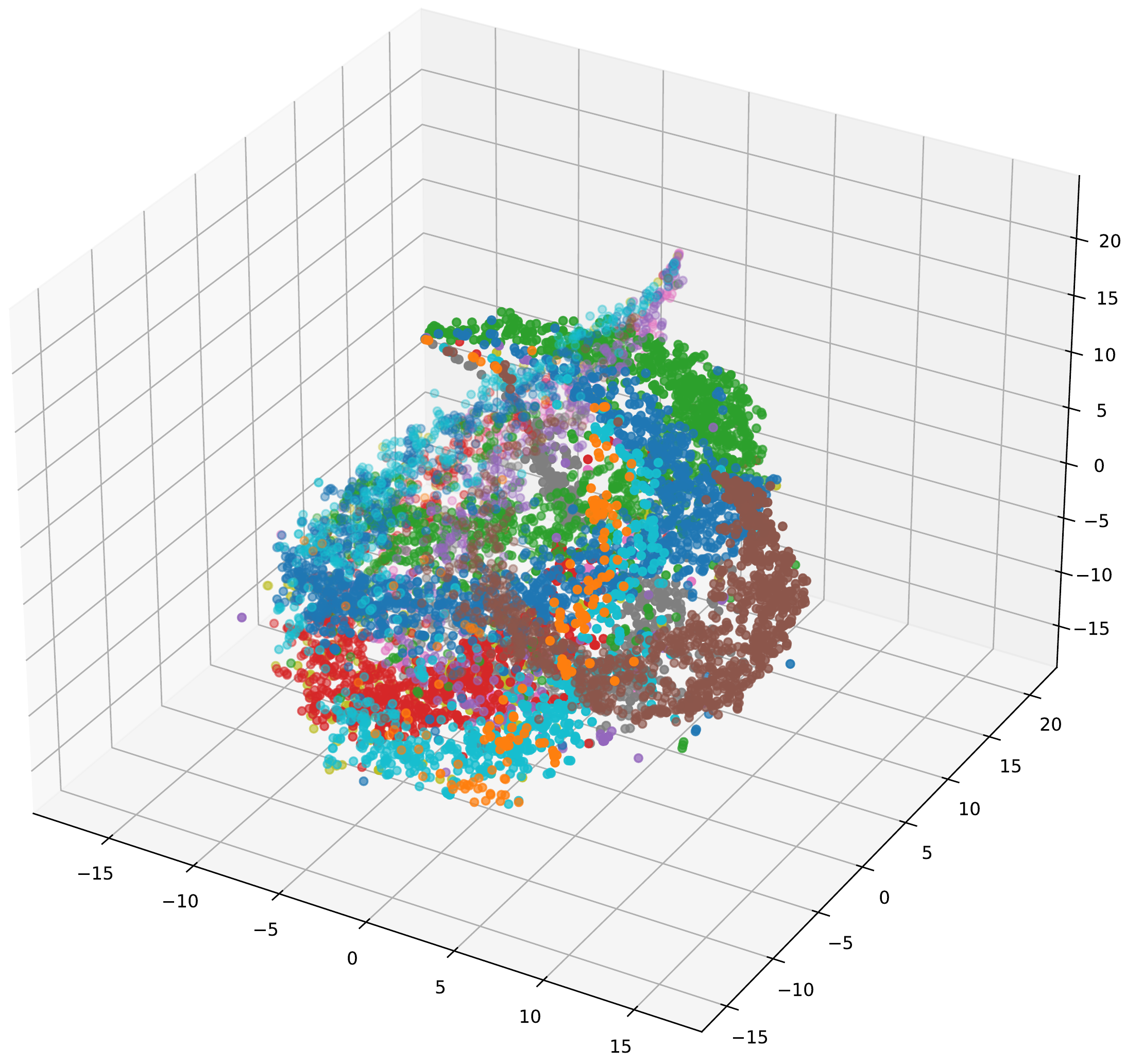}
    \subcaption{$uns$}
  
  \end{subfigure}
  \hfill
  \begin{subfigure}{0.49\linewidth}
  \centering
    \includegraphics[width=4cm]{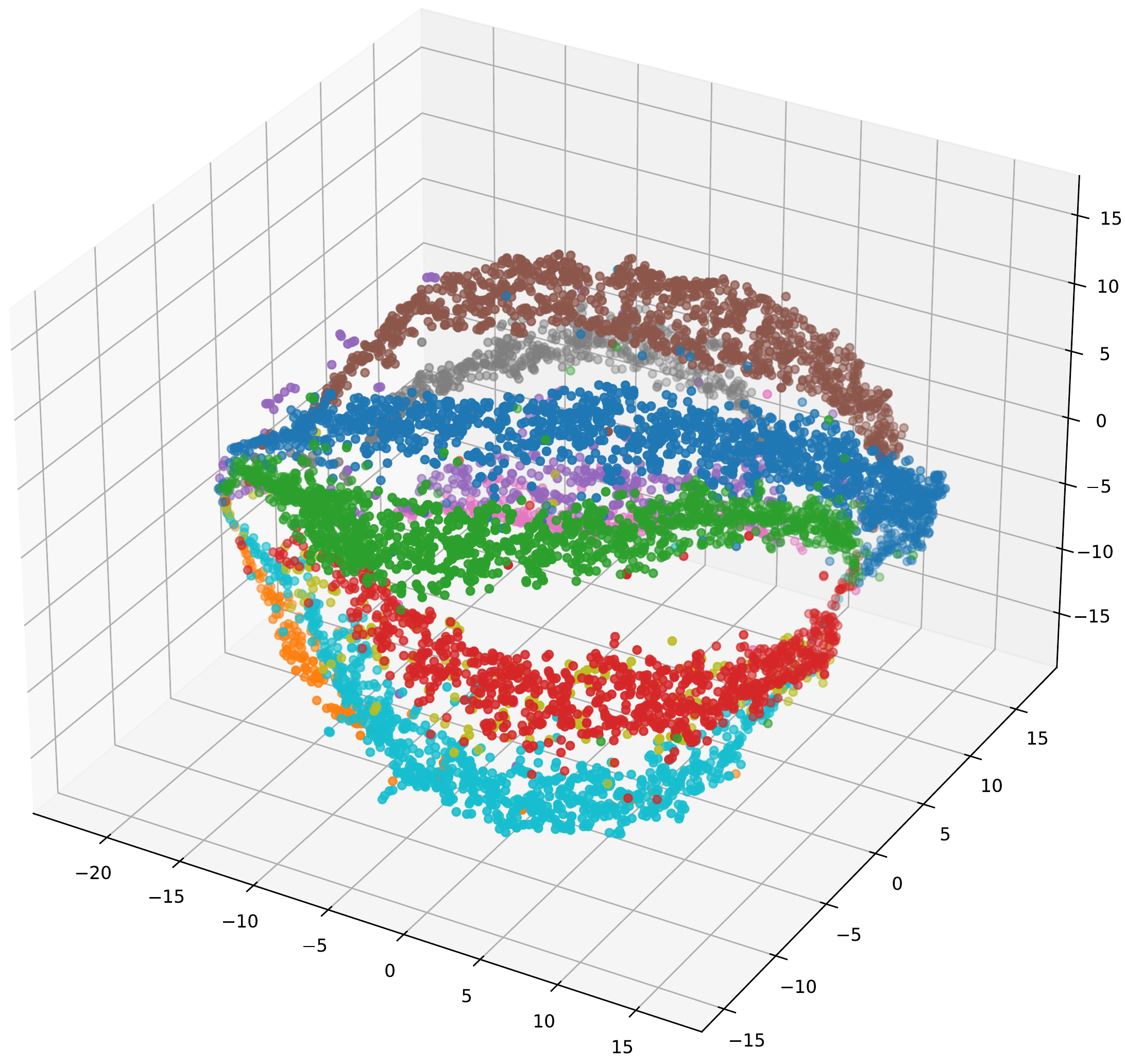}
    \subcaption{$mat$}
  
  \end{subfigure}
  
    \begin{subfigure}{0.49\linewidth}
    \centering
  \includegraphics[width=4cm]{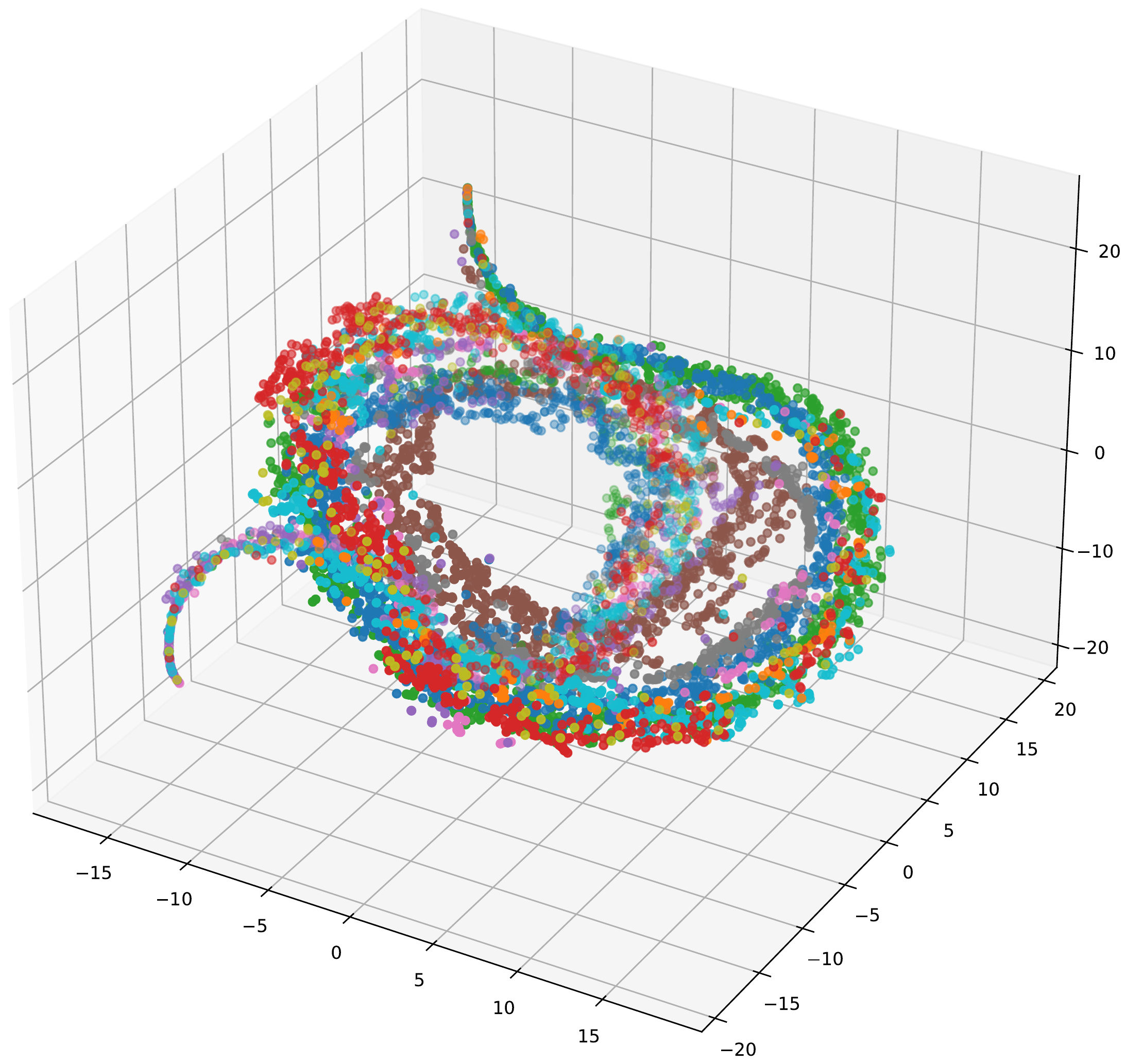}
    \subcaption{$cs$}
   
  \end{subfigure}
  \hfill
  \begin{subfigure}{0.49\linewidth}
  \centering
  \includegraphics[width=4cm]{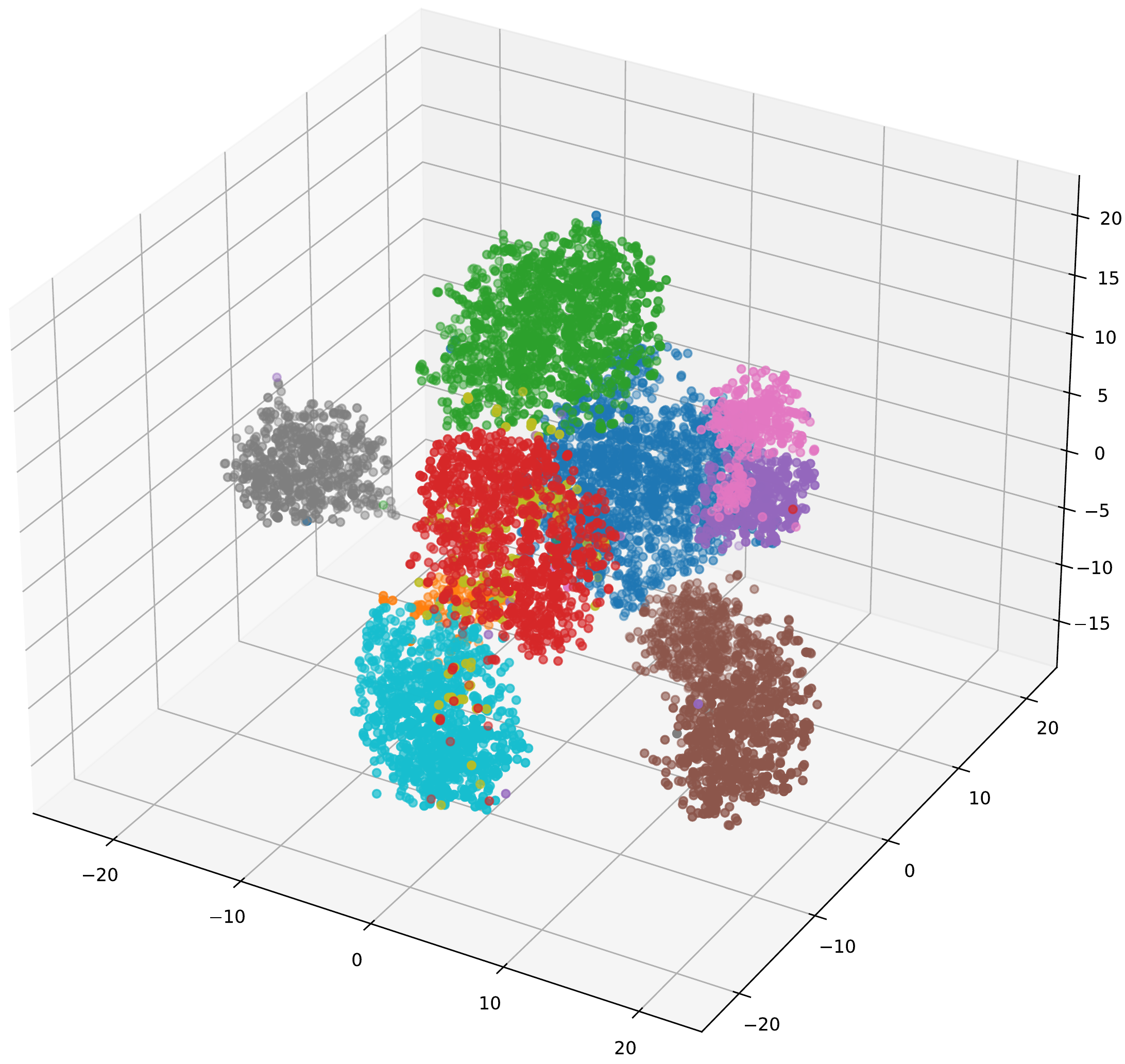}
    \subcaption{$cu$}
 
  \end{subfigure}
  \caption{t-SNE visualization of $uns$, $mat$, $cs$, and $cu$ on the AWA2 dataset.}
  \label{fig:vis}
\end{figure}

\begin{figure} [t]
  \centering
  \begin{subfigure}{0.49\linewidth}
  \includegraphics[width=4cm]{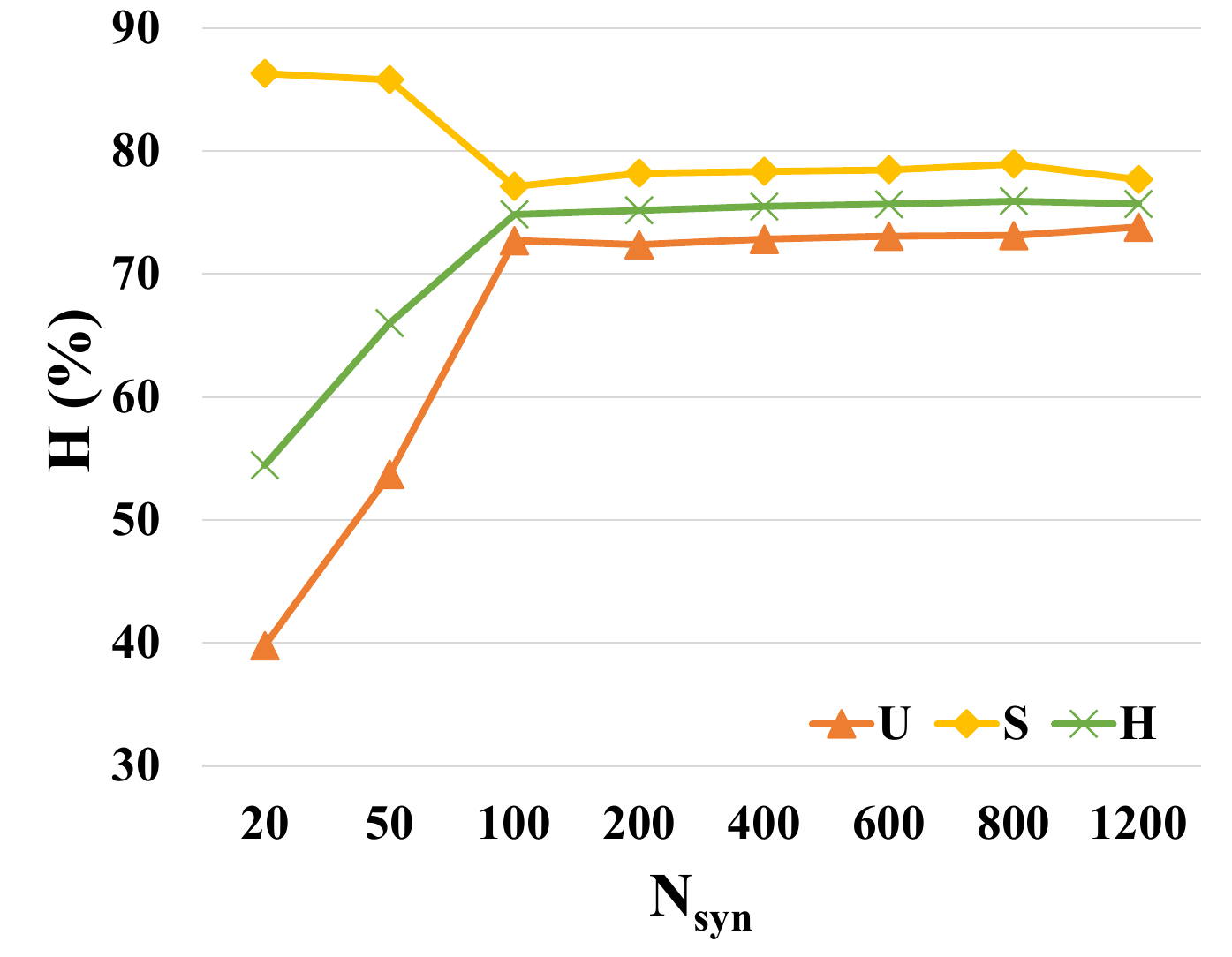}
    \caption{CUB}
    
  \end{subfigure}
  \hfill
    \begin{subfigure}{0.49\linewidth}
  \includegraphics[width=4cm]{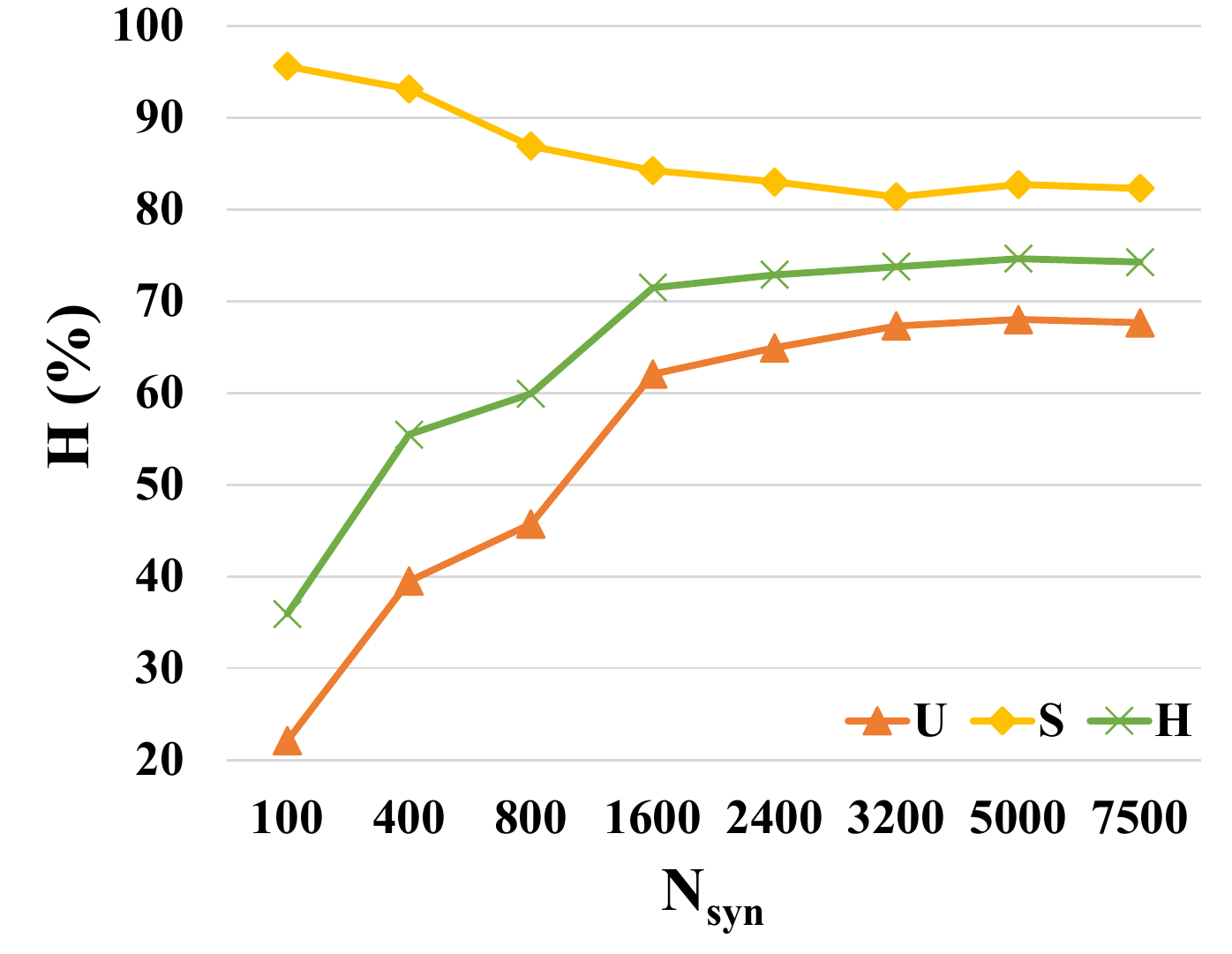}
    \caption{AWA}
 
  \end{subfigure}

    \begin{subfigure}{0.49\linewidth}
  \includegraphics[width=4cm]{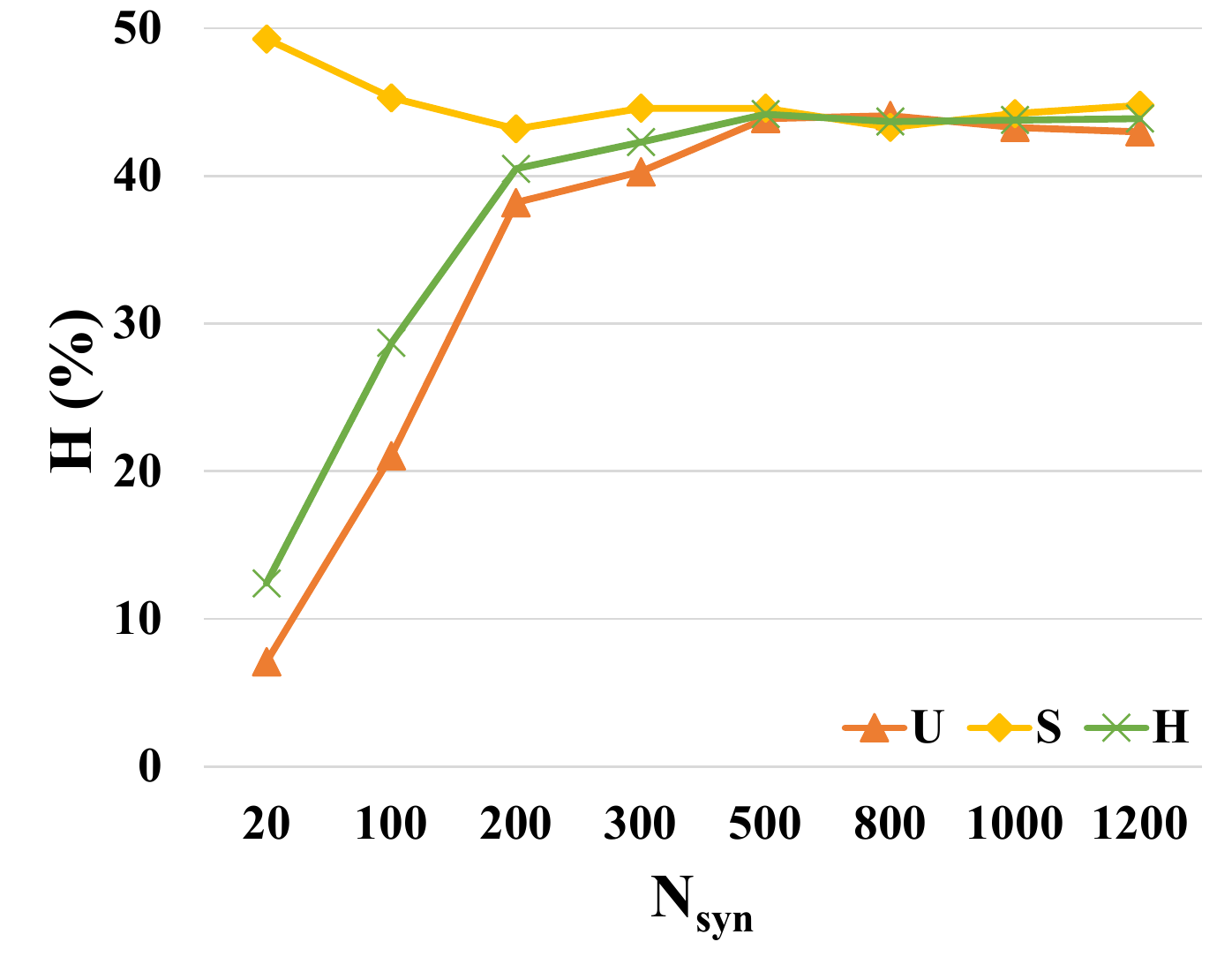}
    \caption{SUN}

  \end{subfigure}
  \hfill
  \begin{subfigure}{0.49\linewidth}
  \includegraphics[width=4cm]{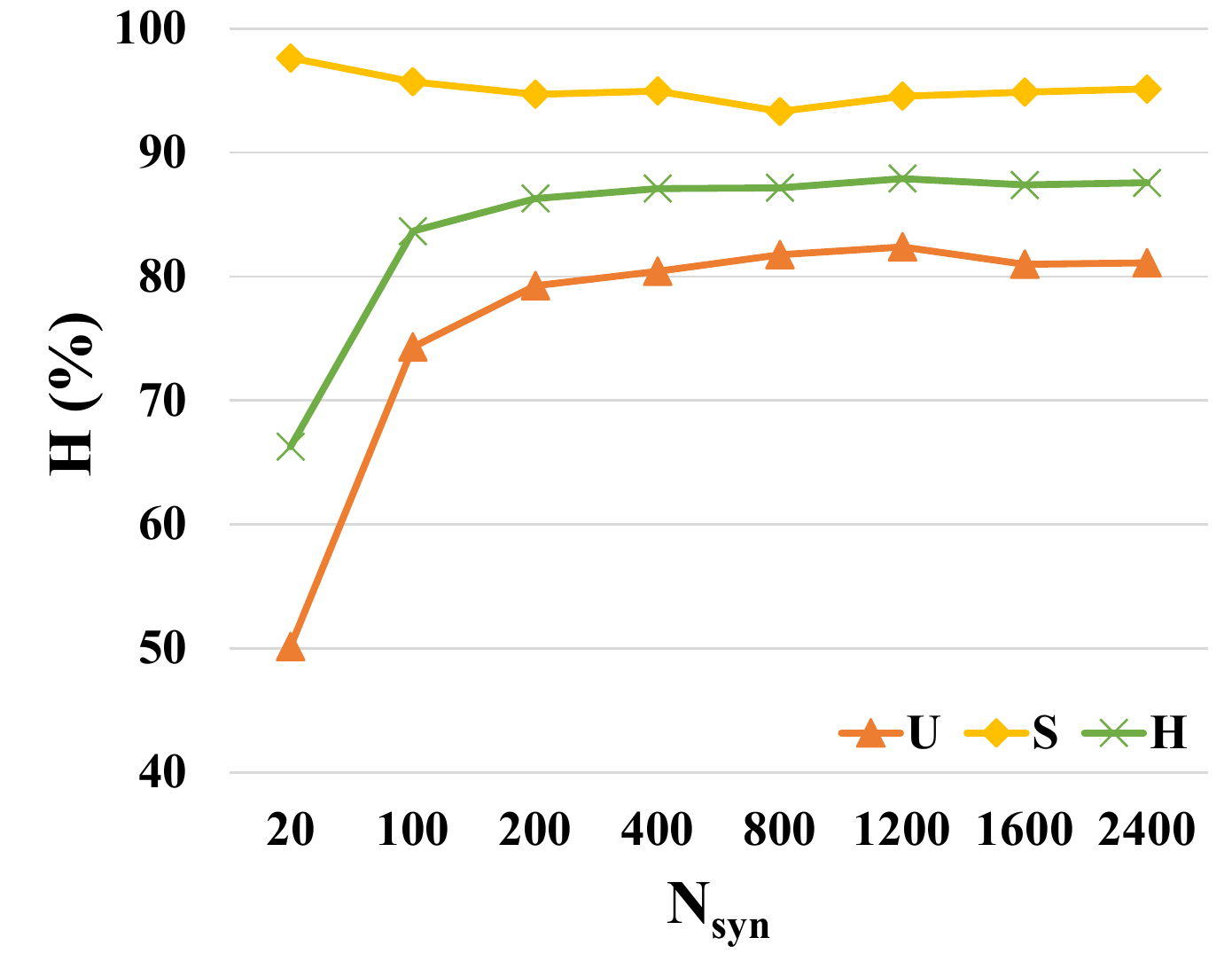}
    \caption{FLO}
   
  \end{subfigure}
  \caption{Evaluation of the performance impact of the number of synthetic unseen classes $N_{syn}$.}
  \label{fig:nsyn}
\end{figure}

\begin{figure*}[htbp]
  \centering
   \includegraphics[width=0.99\linewidth]{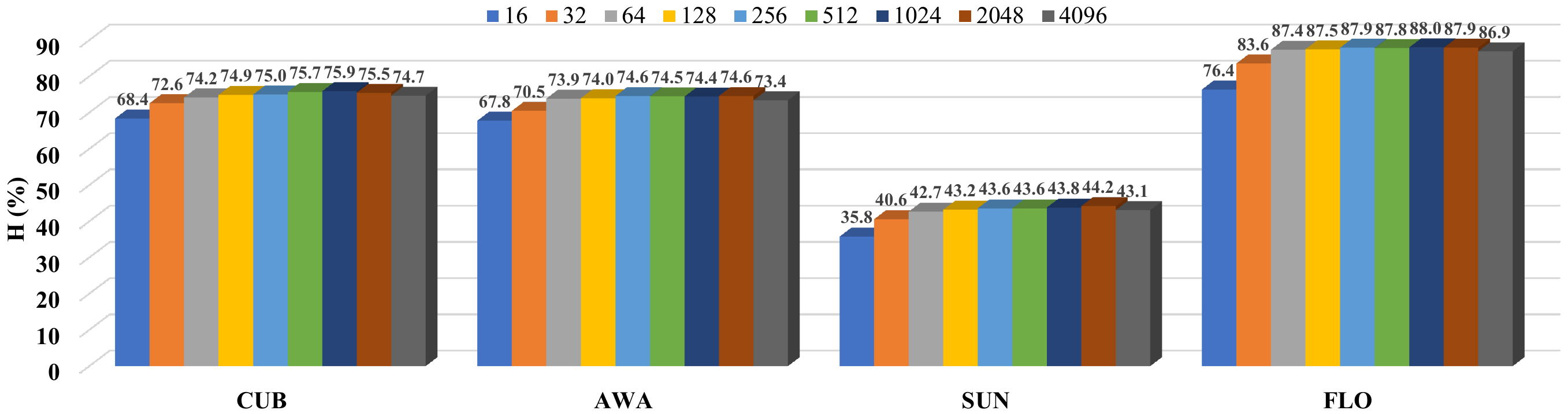}
   \caption{Analysis of the impact of batch size on our proposed method.}
   \label{fig:batchsize}
\end{figure*}

To further verify whether our model is successfully disentangling and disentangling can help to improve the accuracy of discrimination, we conducted a dimensionality reduction visualization experiment by t-SNE. \Cref{fig:vis} respectively shows the visual clustering results of the semantic-unspecific($uns$), semantic-matched($mat=cs+cu$), class-shared($cs$) and class-unique($cu$) vectors are visualized  Specifically, our input is all samples of 10 unseen classes of AWA2. It can be clearly observed that $cu$ clustering is better than uns, in contrast, uns is slightly messy. This shows that $cu$ has strong discriminative information. Intuitively, we think that the semantic-unspecific information displayed by uns is detrimental to the classification accuracy, while $cu$ is the opposite. $cs$ can observe the similarity between classes in the visualization results, but there are also some discriminative manifestations, which is consistent with our idea: $cu$ has basic information, After combining $cs$ and $cu$, mat can use t-SNE clustering well, with higher discriminability than $cs$ and stronger reliability than $cu$. This is why we decided to use mat to train the classifier.

\subsection{Hyper-Parameter Analysis}
\noindent\textbf{Number of synthetic unseen samples $N_{syn}$.} In the \Cref{fig:nsyn}, we evaluated the impact of the number of synthetic unseen samples$N_{syn}$ on our proposed method. It can be seen that as $N_{syn}$ increases, U and H first increase and then basically remain flat, and S first decreases and then basically remain flat. This phenomenon shows that when $N_{syn}$ increases to a certain number, the bias problem between seen and unseen class will be alleviated to some extent. Keeping it flat in the back illustrates that our method also has a stable mitigation effect on the bias problem.

\noindent\textbf{Batch size $B$.} Generally speaking, contrastive learning requires a relatively large batch size. For this key parameter, we conducted experiments on 4 data sets to find an appropriate batch size. \Cref{fig:batchsize} shows that we get the optimal results on CUB, AWA, SUN, FLO When the batch size is set to $\left\{1024, 256, 2048, 1024\right\}$, and the results of setting a moderate value for the four datasets do not differ much, which shows that our method is not very sensitive to $B$.
\begin{figure}[htbp]
  \centering
   \includegraphics[width=0.99\linewidth]{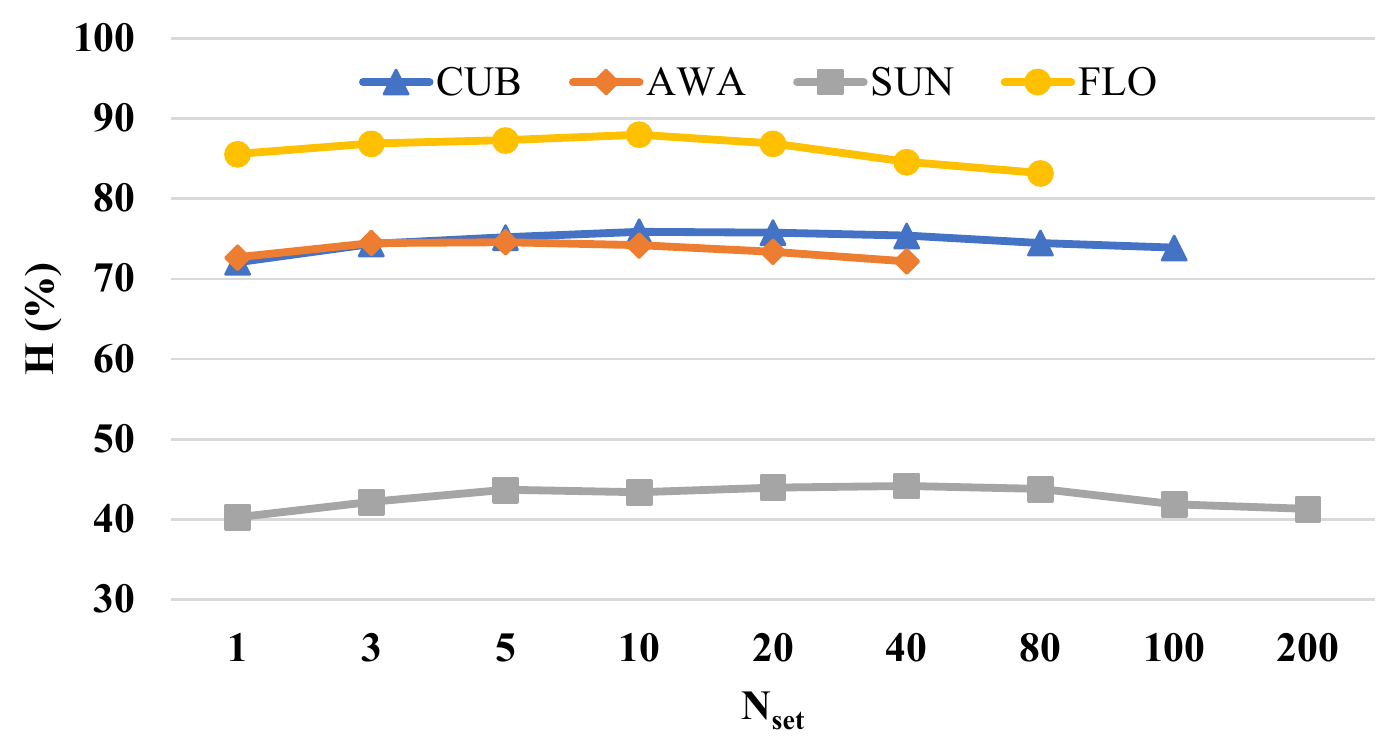}
   \caption{Analysis of the impact of the number of sets on our proposed method. Note that the missing part is because the number of seen classes is less than the number of sets.}
   \label{fig:nset}
\end{figure}
\noindent\textbf{Number of set $N_{set}$.} Our method needs to cluster a batch of samples into $N_{set}$ sets. Obviously, $N_{set}$ has a certain influence on the method. In the experiment, $N_{set}$ is set to $\left\{1, 3, 5, 10, 20, 40, 80, 100, 200\right\}$. Note that the number of seen classes for CUB, AWA, SUN, and FLO are 150, 40, 645, and 82 respectively. There is no experimental result for $N_{set}$ larger than the number of seen classes. \Cref{fig:nset} shows the effect of different $N_{set}$ on four data sets. The results indicate that the performance is best when $N_{set}$ is set to approximately $\frac{1}{10}$ of the seen class.

\section{Conclusion}
We propose a Cluster-based Contrastive Disentangling (CCD) method for GZSL by alleviating semantic gap and domain shift problems. Specifically, we first disentangle clustered visual features into semantic-unspecific and semantic-matched variables. Then, we further disentangle semantic-matched variables into class-shared and class-unique variables. By the introduce of inter-set and inter-class contrastive learning, the semantic-matched variables contain underlying and discriminative information, simultaneously. Our proposed method fully exploits the similarity and discriminability between sets and categories. Experiments on four data sets prove that our proposed method is effective and performs better than other state-of-the-arts.

{\small
\bibliographystyle{ieee_fullname}
\bibliography{egbib}
}

\end{document}